\documentclass[runningheads]{llncs}
\usepackage[T1]{fontenc}
\usepackage{graphicx}
\usepackage{float}
\usepackage{placeins}
\usepackage{booktabs}
\usepackage{amsmath,amssymb}

\usepackage{xcolor}
\usepackage[breaklinks=true,hidelinks]{hyperref}
\usepackage{xr-hyper}
\newcommand{\biblink}[2]{\newblock\href{#1}{#2}}
\makeatletter
\def\@seccntformat#1{\csname the#1\endcsname\quad}
\renewcommand{\@fnsymbol}[1]{\ensuremath{%
  \ifcase#1\or\dagger\or{\dagger\dagger}\or\star\or{\star\star}\or{\star\star\star}\or
  \mathchar"278\or\mathchar"27B\or\|\or**\else\@ctrerr\fi}}
\makeatother
\begin{document}
\title{CARE: A Lightweight Plug-in Gated Correction and Uncertainty-aware Module for Long-term Time Series Forecasting}
\titlerunning{CARE: A Gated Correction and Uncertainty-aware Module for LTSF}
% If the paper title is too long for the running head, you can set
% an abbreviated paper title here
%
\author{Guo Cheng\textsuperscript{1} \and Changlong Lv\textsuperscript{1} \and Jingyi Hou\textsuperscript{1,2,3}\thanks{Corresponding author.}}
\authorrunning{G. Cheng et al.}
% First names are abbreviated in the running head.
% If there are more than two authors, et al. is used.
%
\institute{
\textsuperscript{1} School of Artificial Intelligence, University of Science and Technology Beijing, Beijing 100083, China\\
\textsuperscript{2} Institute of Artificial Intelligence, University of Science and Technology Beijing, Beijing 100083, China\\
\textsuperscript{3} Key Laboratory of Intelligent Bionic Unmanned Systems, Ministry of Education, University of Science and Technology Beijing, Beijing 100083, China\\
\email{\{u202343375, u202343384\}@xs.ustb.edu.cn, houjingyi@ustb.edu.cn}
}
\maketitle % typeset the header of the contribution
\begin{abstract}
Multivariate long-horizon forecasting is critical to electricity load scheduling and traffic flow management, and to financial risk control. Existing deterministic backbones output a single trajectory, masking heterogeneous prediction difficulty across horizons and channels and providing no localized reliability signal.
We present CARE (Corrective branch with Aligned context and Relative-error Estimation), a lightweight plug-in that enhances any deterministic forecaster without architectural redesign. Operating in parallel with the base model, CARE resamples historical context to match the forecast horizon, learns residual correction patterns from this aligned history, and applies scale-aware bounded updates modulated by per-coordinate sigmoid risk gates. A multi-objective loss jointly optimizes forecast accuracy, residual tracking, risk alignment, and base-model anchoring.
Across eight benchmarks with three representative backbones, CARE improves accuracy with marginal parameter and latency overhead. Its risk gates reliably identify high-error regions: on Weather, the highest-gate tertile exhibits nearly four times the error of the lowest-gate tertile, offering planners an interpretable per-step trust signal.
Code is available at \url{https://github.com/CG-BNYC/CARE}.

\end{abstract}
\section{Introduction}
\begingroup
\setlength{\parskip}{0pt plus 0pt minus 0pt}
Multivariate long-horizon time series forecasting is critical to electricity load scheduling and traffic flow management~\cite{lai2018modeling}, and to financial risk control~\cite{benidis2020neural}. Recent backbones---PatchTST~\cite{nie2023patchtst}, Mamba~\cite{gu2024mamba}, DLinear~\cite{zeng2023dlinear}, and their variants---achieve strong average accuracy by learning rich temporal patterns from long histories. Yet these models share a fundamental limitation: they compress historical information into a single deterministic trajectory, treating every future time step and every variable as equally predictable. This homogenization masks the heterogeneous error structure inherent in real-world data, where some horizons are inherently noisier and some channels more volatile. Consequently, planners receive no per-step, per-variable reliability signal to hedge against high-uncertainty segments, even when aggregate error metrics appear low.

Several strategies have emerged to address this gap, but each carries significant drawbacks. Probabilistic forecasters such as DeepAR~\cite{salinas2020deepar} and TFT~\cite{lim2021temporal} are built around distributional outputs rather than point predictions alone, so porting comparable uncertainty signals to an existing deterministic backbone still requires substantial architectural redesign. Post-hoc conformal calibration operates on frozen predictions~\cite{vovk2005algorithmic,romano2019conformalized,angelopoulos2024gentle}, unable to co-evolve representations with the base model. Stacked iterative refinement~\cite{sen2019think} adds sequential inference cost~\cite{lee2018deterministic}, undermining the efficiency that makes deterministic backbones attractive. What remains elusive is a portable, lightweight enhancement that preserves an existing point forecaster while providing localized correction and uncertainty indication.

Naively attaching a correction branch to a base forecaster faces three barriers. First, lookback length and forecast horizon typically differ~\cite{nie2023patchtst,wu2021autoformer}, causing length--semantic misalignment: a convolutional kernel at position $i$ in the history does not correspond to step $t$ in the future. Second, feeding the base forecast into the branch risks re-fitting or amplifying the preliminary output rather than learning genuine residual patterns. Third, unconstrained corrections or unsupervised gates~\cite{sen2019think,wang2023micn,han2024softs} can destabilize training---either overturning a competent base model or collapsing to near-zero updates.

We propose CARE (Corrective branch with Aligned context and Relative-error Estimation), a lightweight parallel plug-in trained end-to-end with any deterministic forecaster. CARE learns residual corrections from temporally aligned historical context and modulates their contribution through a per-step risk gate. Co-trained end-to-end with the base model, it yields refined predictions and an interpretable reliability signal without architectural redesign or heavy inference cost. Across eight standard benchmarks~\cite{nie2023patchtst,zhou2021informer,wu2021autoformer} with PatchTST, Mamba, and DLinear backbones~\cite{nie2023patchtst,gu2024mamba,zeng2023dlinear}, CARE delivers consistent accuracy improvements with marginal parameter and latency overhead. Its risk gates reliably identify high-error regions: on Weather, the highest-gate tertile exhibits nearly four times the absolute error of the lowest-gate tertile, confirming the gate as a meaningful indicator of local prediction difficulty.

\subsubsection{Contributions}
\begin{itemize}
\item We transform the prediction-correction framework from a post-hoc or stacked structure into a parallel branch internal to the deterministic forecaster, enabling end-to-end co-training where the correction operator evolves jointly with the backbone rather than being tuned in isolation.
\item We propose CARE, a lightweight plug-in that generates horizon-semantically compatible input through length-aligned historical resampling, applies gated residual correction with scale-aware bounded updates, and employs multi-objective joint training to ensure the correction branch stably enhances the base forecaster.
\item We validate CARE's portability across three representative backbones and eight standard benchmarks. Results demonstrate consistent and stable accuracy gains with minimal overhead in both parameter count and inference latency.
\end{itemize}
\endgroup

\section{Related Work}
\label{secrelated}

\subsection{Deep Models for Long-Horizon Forecasting}
Recurrent~\cite{hochreiter1997long,cho2014empirical,lai2018modeling} and convolutional~\cite{bai2018empirical,lai2018modeling} encoders were early workhorses for sequence
prediction,
while recent systems scale to long lookbacks with Transformer-style attention~\cite{vaswani2017attention,li2019enhancing,wu2021autoformer,zhou2022fedformer} and improved complexity
tricks~\cite{zhou2021informer,liu2022pyraformer}.
Decomposition, frequency enhancement, and multi-scale temporal modeling~\cite{wu2021autoformer,zhou2022fedformer,wu2023timesnet,liu2022scinet,wang2023micn} further specialize Transformers to non-stationary real-world
series~\cite{liu2022nonstationary}.
Complementary linear~\cite{zeng2023dlinear}, basis-expansion~\cite{oreshkin2020nbeats,challu2023nhits}, and MLP-mixer~\cite{chen2023tsmixer,tolstikhin2021mlpmixer} designs demonstrate that simple
decoders can be surprisingly strong on standard
benchmarks,
motivating hybrid or post-processing strategies rather than ever-larger
monolithic heads~\cite{benidis2020neural}. Patching and channel-independent
encoding underpin strong point backbones such as PatchTST~\cite{nie2023patchtst},
Mamba~\cite{gu2024mamba}, and DLinear~\cite{zeng2023dlinear}, which we adopt as
representative trunks in our experiments.

\subsection{Prediction-Correction Frameworks}
Prediction-correction appears wherever a provisional estimate is updated after new
evidence, including linear and nonlinear state-space filtering~\cite{kalman1960new}, sequential Monte
Carlo updates~\cite{doucet2011tutorial}, receding-horizon control, and iterative non-autoregressive decoding
in NLP~\cite{lee2018deterministic}. Contemporary
deep multivariate forecasters nonetheless usually compress history into a single
forward map. When a second stage exists, it is typically stacked, calibrated, or
fused only after the main predictor is fixed. We instead embed prediction-correction
as a parallel branch trained end-to-end with the base forecaster under the same
point-forecast objective, so the correction operator co-evolves with the trunk rather
than living in an outer loop or a separate post-processing script.

\subsection{Quantifying Predictive Uncertainty in Forecasting}
Dedicated probabilistic~\cite{salinas2020deepar,lim2021temporal} and state-space heads~\cite{rangapuram2018deep} expose uncertainty by architectural
design. Alternatively,
uncertainty can be injected through Bayesian parameterizations~\cite{blundell2015weight}
or deep ensembles~\cite{lakshminarayanan2017simple}, though at significant training or
inference cost. Quantile~\cite{koenker2005quantile} and proper-scoring objectives~\cite{gneiting2007strictly}
recast the learning target itself, while conformal calibration~\cite{vovk2005algorithmic,romano2019conformalized,angelopoulos2024gentle}
wraps spread estimation around an otherwise point-first map. Residual or gated hybrids~\cite{sen2019think,wang2023micn,han2024softs}
nudge a baseline through modular adjustment. We instead keep the correction path inside
the same forward pass and multitask loss as the base forecaster, instantiating this
integration with length-matched inputs, per-coordinate gates, and auxiliary supervision
in Section~\ref{secmethod}.

\section{Method}
\label{secmethod}

\subsection{Problem Setup and Notation}
We study multivariate long-horizon forecasting from a lookback tensor
$x\in\mathbb{R}^{L\times C}$ with $L$ past steps and $C$ channels to the next $T$
steps. Training and inference both use sliding windows over a longer series, and
we stack minibatches along the batch dimension so that
$x\in\mathbb{R}^{B\times L\times C}$ and
$Y,Y^{\mathrm{base}},Y^{\mathrm{true}}\in\mathbb{R}^{B\times T\times C}$, with
subscripts $b,t,c$ denoting batch, horizon, and channel. The base forecaster described in Subsection~\ref{subsec:baseforecaster} is the deterministic map
$f_{\theta}\colon\mathbb{R}^{L\times C}\to\mathbb{R}^{T\times C}$ that yields the
base forecast $Y^{\mathrm{base}}=f_{\theta}(x)$. With parameters $\psi$, CARE
refines $Y^{\mathrm{base}}$ through a gated correction built from length-matched
history to obtain the final prediction $Y$, trained against $Y^{\mathrm{true}}$.

\subsection{Design Principles}
\begin{figure}[t]
\centering
\includegraphics[width=\linewidth]{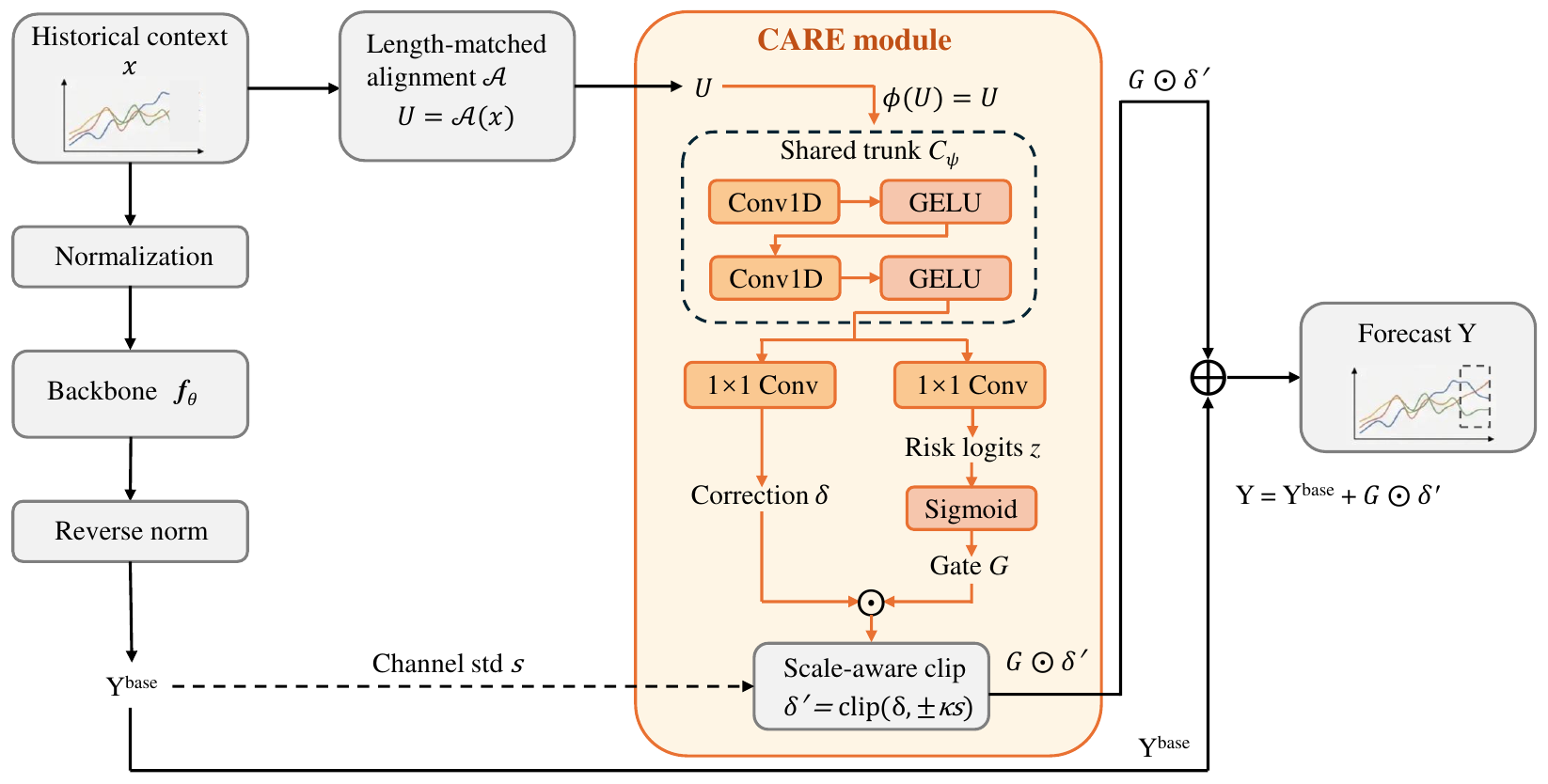}
\caption{Overview of the CARE plug-in.}
\label{fig:care}
\end{figure}

CARE follows four principles.
\textit{P1:} Let the base forecaster remain the main seat of temporal modeling and let the dominant forecasting loss continue to strengthen $f_{\theta}$, with CARE focusing on residual structure the base forecaster tends to miss.
\textit{P2:} Feed the plug-in only length-matched raw history at its first stage, withhold an immediate copy of the preliminary forecast there, and whenever the encoder lookback length differs from the forecast horizon reshape or resample what the plug-in sees so that its local operators are indexed compatibly with the future window.
\textit{P3:} Keep corrections scale-aware and tightly bounded relative to the typical spread of the preliminary trajectory so updates stay selective rather than explosive.
\textit{P4:} Augment the headline forecasting error with mild auxiliary terms that teach the adjustment and gate behaviors, plus a small anchor on the base-forecaster output, so the two pathways co-train without CARE collapsing into a wholesale bypass of $f_{\theta}$.
The remainder of this section details the base forecaster, the CARE architecture, and the joint training objective.
Figure~\ref{fig:care} summarizes the overall data flow.

\subsection{Base Forecaster and Base Prediction}
\label{subsec:baseforecaster}
We use \emph{base forecaster} for the complete deterministic branch $f_{\theta}$
that outputs $Y^{\mathrm{base}}$---not the encoder alone. The map includes the
forecasting backbone core, the prediction head (e.g., the PatchTST flatten head),
and the full ReVIN reversible instance normalization pipeline from affine
instance normalization through inverse denormalization~\cite{kim2022revin}; CARE is
outside $f_{\theta}$.
The trunk is any long-horizon model of this form that maps encoded history $x$ to a
preliminary tensor over the prediction horizon. We denote the resulting
\emph{base} forecast by
\begin{equation}
 Y^{\mathrm{base}} = f_{\theta}(x)\in\mathbb{R}^{B\times T\times C}.
\end{equation}
The functional module $f_{\theta}$ can be implemented by various publicly available architectures. In our experiments, we mainly adopted three representative instantiated models, namely PatchTST, Mamba, and DLinear~\cite{nie2023patchtst,gu2024mamba,zeng2023dlinear}. For the complete block diagrams and training details of each base forecaster, please refer to the corresponding papers.
All subsequent definitions of CARE assume that the branch processes historical
information in parallel with $Y^{\mathrm{base}}$ and refines the preliminary output
within the same tensor space as $Y^{\mathrm{base}}$. This design enables the
base forecaster to be freely replaced.

\subsection{CARE Architecture}
\subsubsection{Length-Matched Input Context}
CARE consumes history, but naively passing the raw length-$L$ tensor into
a module whose output must live on the $T$-step forecast grid mixes two
different temporal semantics, so kernel position along $L$ does not correspond to
\emph{step $t$ of the future}. We therefore ask CARE to operate on a
\emph{length-matched} view of the past, namely a length-$T$ signal per channel obtained
by resampling the same information $x$ carries, yet indexed compatibly with the
prediction tensor. Concretely, we apply a per-channel resampling operator
$\mathcal{A}$. If $L\neq T$, for each $(b,c)$ we perform 1-D linear interpolation
of $x_{b,1,\ldots,L,c}$ onto $T$ uniformly spaced grid points, and if $L{=}T$, $\mathcal{A}$
is the identity. Linear interpolation is a smooth, parameter-free alignment, and it
favors stability over introducing extra learnable warping layers that could
overfit small benchmarks. The aligned history is
\begin{equation}
 U=\mathcal{A}(x)\in\mathbb{R}^{B\times T\times C}.
\end{equation}
We set the CARE feature to the aligned history itself,
\begin{equation}
 \label{eq:care-input}
 \Phi(U)=U,
\end{equation}
consequently the first convolutional layer of CARE has $C$ input
channels, forcing CARE to infer correction cues purely from historical dynamics
resampled to length $T$.
We also evaluated several alternative input designs for CARE in experiments:
$Y^{\mathrm{base}}$ alone, the concatenation of $Y^{\mathrm{base}}$ and $U$, and
$Y^{\mathrm{base}}-U$.
Each variant performed worse than the input in Eq.~\eqref{eq:care-input}, so we
adopt $\Phi(U){=}U$ throughout.

\subsubsection{Convolutional Critic}
CARE receives $\Phi(U)\in\mathbb{R}^{B\times T\times F}$, where $F$ is the
number of input feature channels. The input has $F{=}C$ feature channels in all
settings considered above, except when $\Phi(U)=[Y^{\mathrm{base}},U]$, where
$F{=}2C$.
Before convolution, we permute the tensor to $V\in\mathbb{R}^{B\times F\times T}$
so that each of the $F$ channels is a 1-D signal over the $T$ forecast steps.
Convolution is
chosen because it implements \emph{local} mixing along the horizon with few
parameters, and nearby $t$ often share error structure (e.g., smooth multi-step
drift in the base forecast), and small kernels let CARE exploit that
regularity without building another full attention mechanism. A shallow
convolutional trunk $\mathcal{C}_{\psi}$ applies two temporal 1-D convolution
layers with zero padding chosen so the temporal length remains $T$, using GELU
activation. The convolutional trunk is shared. Two $1{\times}1$ linear readouts
split the shared representation into a correction branch and a risk branch, with
$\sigma$ the elementwise logistic sigmoid applied to the risk logits,
\begin{equation}
 \label{eq:care-heads}
 (\delta, z) = \mathcal{C}_{\psi}(\Phi(U)),
\end{equation}
where $\delta,z\in\mathbb{R}^{B\times T\times C}$.
For each $(b,t,c)$, $\delta_{b,t,c}$
proposes how much to adjust the base forecast at that $(t,c)$, and $z_{b,t,c}$
is the corresponding gate logit. $G_{b,t,c}=\sigma(z_{b,t,c})\in(0,1)$ converts
it into a trust weight that scales the correction before merging with
$Y^{\mathrm{base}}$.
Decoupling $\delta$ and $z$ at the readout layer
is important, so one branch answers \emph{how much we should move} and the other
\emph{whether we are confident enough to move there}, instead of entangling both into a
single unconstrained vector.

\subsubsection{Scale-Aware Clipping}
Large unconstrained $\delta$ can destabilize training and amplify forecaster
noise, especially when the base trajectory is already smooth along $t$. For each
$(b,c)$, let $\mu_{b,c}=\frac{1}{T}\sum_{t=1}^{T}
Y^{\mathrm{base}}_{b,t,c}$ and define the time-wise standard deviation with
population divisor $T$ and a small stabilizer $\varepsilon>0$, with
\begin{equation}
 s_{b,c} =
 \sqrt{\frac{1}{T}\sum_{t=1}^{T}\bigl(Y^{\mathrm{base}}_{b,t,c}-\mu_{b,c}\bigr)^2+\varepsilon}.
\end{equation}
We broadcast $s$ over $t$ and map each entry of $\delta$ into $[-\kappa s, +\kappa s]$,
yielding the capped correction $\delta'$,
with hyperparameter $\kappa{>}0$.
In our experiments we evaluate several choices of $\kappa$; see Appendix
Table~\ref{tab:kappa} for the sweep.
Intuitively, channels
whose base trajectory is flat admit only small corrections, whereas volatile
channels may receive larger updates. This cap is not a probabilistic uncertainty
calibration in the Bayesian sense, and it is a training-time and run-time stability
prior that nonetheless interacts well with the risk loss on $G$, because large
$\delta$ is discouraged where $G$ should already shrink the update.

\subsubsection{Gated Merge}
Training and inference both use the merged output, again with $\odot$ the
elementwise product,
\begin{equation}
 \label{eq:finalpred}
 Y = Y^{\mathrm{base}} + G \odot \delta'.
\end{equation}
The gate $G$ soft-selects where the residual is applied, so near-zero entries
preserve the trunk forecast, while entries near one allow CARE to inject a
bounded correction.
The advantage of this multiplicative interaction is its smoothness: compared
with discrete hard masks, continuous gate values provide more stable gradients
during back-propagation, so the base forecaster and correction branches can be
co-optimized jointly.
From an interpretability perspective, $G$ forms a continuous trust map over
the spatiotemporal domain $(t,c)$, quantifying how strongly the model trusts and
calibrates its forecast at each time step and variable.

\subsection{Multi-Objective Training}
A single squared error on $Y$ would technically be sufficient to make the
network trainable, but it provides weak direct supervision for \emph{how} $\delta$
and $G$ should divide labor, since many $(\delta,G)$ pairs yield similar $Y$. We
therefore expose additional training signals that make the internal roles of CARE interpretable and reduce pathological solutions where the gate is
ignored or always saturated.

\subsubsection{Prediction Loss}
The main signal remains regression on the final prediction, with
\begin{equation}
 \mathcal{L}_{\mathrm{pred}} =
 \frac{1}{BTC}\bigl\lVert Y - Y^{\mathrm{true}}\bigr\rVert_F^2.
\end{equation}

\subsubsection{Residual Loss}
Let the true residual of the base forecast be
$E=Y^{\mathrm{true}}-Y^{\mathrm{base}}$.
We set $E^{\mathrm{sg}}=\mathrm{stopgrad}(E)$ so that in $\mathcal{L}_{\delta}$
the target $E$ is fixed with respect to network parameters, with gradients
updating $\delta$ only and not back-propagating through
$Y^{\mathrm{true}}-Y^{\mathrm{base}}$ into $Y^{\mathrm{base}}$.
Otherwise the trunk could shrink $E$ by changing its own prediction without
$\delta$ actually tracking the current error pattern.
Define
\begin{equation}
 \mathcal{L}_{\delta} =
 \frac{1}{BTC}\bigl\lVert \delta - E^{\mathrm{sg}}\bigr\rVert_F^2.
\end{equation}
Thus $\mathcal{L}_{\delta}$ trains $\delta$ to characterize the base forecaster's
present error structure.

\subsubsection{Risk Loss}
For each batch $b$, define the mean scale
$n_b=\frac{1}{TC}\sum_{t=1}^{T}\sum_{c=1}^{C}\lvert E_{b,t,c}\rvert$.
We form a target risk map in $[0,1]$ by normalizing $\lvert E_{b,t,c}\rvert$ with
$n_b$ and capping values above one at one,
\begin{equation}
 R_{b,t,c}=\min\!\left(1,\frac{\lvert E_{b,t,c}\rvert}{n_b}\right).
\end{equation}
The normalization by $n_b$ makes $R$ comparable across samples with different
overall volatility, so a large absolute error is \emph{risky} relative to the
typical error magnitude in that instance, not in absolute physical units alone.
The risk loss aligns gates to where the base model errs more (relative to the
average absolute error of the sample), with
\begin{equation}
 \mathcal{L}_{\mathrm{risk}} =
 \frac{1}{BTC}\sum_{b,t,c}\bigl(G_{b,t,c}-R_{b,t,c}\bigr)^2.
\end{equation}
Thus $G$ is not only a free nonlinearity for improving MSE, but it is explicitly
pulled toward a human-interpretable notion of relative mistake size.

\subsubsection{Anchor Loss}
To prevent CARE from drifting into solutions that ignore the forecaster
entirely, we add a mild anchor on the base-forecaster output, with
\begin{equation}
 \mathcal{L}_{\mathrm{anchor}} =
 \frac{1}{BTC}\bigl\lVert Y^{\mathrm{base}} - Y^{\mathrm{true}}\bigr\rVert_F^2.
\end{equation}
This term does not compete numerically with $\mathcal{L}_{\mathrm{pred}}$ for
the same optimum. Instead, it regularizes the joint dynamics so that shrinking
$Y^{\mathrm{base}}$ to zero is never an easy way to reduce error when CARE
overfits.

\subsubsection{Combined Objective}
The scalar optimized per step is
\begin{equation}
 \label{eq:combined-loss}
 \mathcal{L}=\mathcal{L}_{\mathrm{pred}}
 +\mathcal{L}_{\delta}
 +\mathcal{L}_{\mathrm{risk}}
 +\mathcal{L}_{\mathrm{anchor}}.
\end{equation}

\section{Experiments}
\label{secexp}
\FloatBarrier

\subsection{Experimental Setup}
\label{subsec:expsetup}

\subsubsection{Datasets}
We evaluate on standard multivariate long-horizon benchmarks---ETT(ETTh1, ETTh2,
ETTm1, ETTm2), Electricity, Traffic, National ILI(ILI), and
Weather~\cite{zhou2021informer,wu2021autoformer,lai2018modeling}.
The forecasting task uses shared
fixed-length multivariate history as input and predicts each channel
over that benchmark's horizon. Following the usual protocol, horizons are
are $\{96,192,336,720\}$ for ETT, Weather, Traffic, and Electricity
and $\{24,36,48,60\}$ for ILI.

\subsubsection{Base Forecaster}
\label{subsec:expbaseforecaster}
Throughout the experiments we instantiate the deterministic map $f_{\theta}$ from
Section~\ref{subsec:baseforecaster} with three representative long-horizon backbones,
PatchTST, Mamba, and DLinear~\cite{nie2023patchtst,gu2024mamba,zeng2023dlinear}.
Each run uses the full forecasting pipeline of the chosen backbone, including its
encoder, horizon-specific prediction head, and channel-independent encoding with
reversible instance normalization (RevIN)~\cite{kim2022revin} inside $f_{\theta}$.
Architectural details and backbone-specific training recipes are taken from the
original papers.

\subsubsection{Implementation Details}
\label{subsec:impl}
We follow the data splits and preprocessing protocol of PatchTST~\cite{nie2023patchtst}.
CARE is attached with the length-matched input $\Phi(U){=}U$ unless an input-mode study
(Section~\ref{subsec:input}) explicitly states otherwise.
CARE uses a two-layer temporal 1-D convolutional trunk with kernel size 3, hidden width 64.
Clipping factor $\kappa{=}0.1$ unless ablated in Section~\ref{subsec:ablation}.
The joint objective in Eq.~\eqref{eq:combined-loss} is used for the full model while
ablations drop individual terms as indicated in Section~\ref{subsec:ablation}.
All models are trained end to end for 100 epochs with Adam ($\mathrm{lr}{=}10^{-4}$,
batch size 8 for Electricity, Traffic, and 128 for other datasets) on NVIDIA GeForce RTX 4090. Early stopping uses validation loss.

\subsubsection{Metrics}
We report test-set mean squared error (MSE) and mean absolute error (MAE), each
averaged over all future time steps and variates.
Lower values indicate better forecasts.
Every configuration is repeated with five random seeds. We report the mean test metric per dataset over all horizons.
Full results are available in Appendix.

\subsection{Gains Across Base Forecasters}
\label{subsec:crossbase}
Table~\ref{tab:crossbase} reports test errors on all eight benchmarks, averaged over
prediction horizons per dataset.
Panel~(a) lists MSE and panel~(b) lists MAE.
\emph{Base} is the base forecaster $f_{\theta}$ only and \emph{+CARE} attaches the gated
correction branch.

\begin{table}[t]
\caption{Test MSE and MAE across base forecasters, averaged on prediction horizons.}
\label{tab:crossbase}
\centering
\small
\setlength{\tabcolsep}{5pt}
\noindent
\begin{minipage}[c]{1.4em}
  \textbf{(a)}\label{tab:crossbase_mse}
\end{minipage}\hspace{0.4em}%
\begin{minipage}[c]{\dimexpr\linewidth-1.8em\relax}
\centering
\begin{tabular}{lcccccc}
\toprule
Dataset &
\multicolumn{2}{c}{PatchTST} &
\multicolumn{2}{c}{Mamba} &
\multicolumn{2}{c}{DLinear} \\
\cmidrule(lr){2-3}\cmidrule(lr){4-5}\cmidrule(lr){6-7}
& Base & +CARE & Base & +CARE & Base & +CARE \\
\midrule
ETTh1 & 0.413 & \textbf{0.404} & 0.455 & \textbf{0.437} & 0.423 & \textbf{0.412} \\
ETTh2 & 0.331 & \textbf{0.322} & 0.349 & \textbf{0.337} & 0.431 & \textbf{0.424} \\
ETTm1 & 0.353 & \textbf{0.347} & 0.355 & \textbf{0.339} & 0.357 & \textbf{0.348} \\
ETTm2 & 0.256 & \textbf{0.249} & 0.260 & \textbf{0.258} & 0.267 & \textbf{0.258} \\
Electricity & 0.159 & \textbf{0.153} & 0.164 & \textbf{0.160} & 0.166 & \textbf{0.160} \\
Traffic & 0.391 & \textbf{0.386} & 0.411 & \textbf{0.402} & 0.434 & \textbf{0.422} \\
ILI & 1.612 & \textbf{1.523} & 1.561 & \textbf{1.433} & 2.169 & \textbf{1.988} \\
Weather & 0.226 & \textbf{0.223} & 0.225 & \textbf{0.223} & 0.246 & \textbf{0.242} \\
\bottomrule
\end{tabular}
\end{minipage}

\vspace{0.7em}

\noindent
\begin{minipage}[c]{1.4em}
  \textbf{(b)}\label{tab:crossbase_mae}
\end{minipage}\hspace{0.4em}%
\begin{minipage}[c]{\dimexpr\linewidth-1.8em\relax}
\centering
\begin{tabular}{lcccccc}
\toprule
Dataset &
\multicolumn{2}{c}{PatchTST} &
\multicolumn{2}{c}{Mamba} &
\multicolumn{2}{c}{DLinear} \\
\cmidrule(lr){2-3}\cmidrule(lr){4-5}\cmidrule(lr){6-7}
& Base & +CARE & Base & +CARE & Base & +CARE \\
\midrule
ETTh1 & 0.434 & \textbf{0.427} & 0.468 & \textbf{0.454} & 0.437 & \textbf{0.431} \\
ETTh2 & 0.381 & \textbf{0.376} & 0.395 & \textbf{0.388} & 0.447 & \textbf{0.443} \\
ETTm1 & 0.382 & \textbf{0.377} & 0.383 & \textbf{0.376} & 0.379 & \textbf{0.374} \\
ETTm2 & 0.317 & \textbf{0.311} & 0.322 & \textbf{0.318} & 0.332 & \textbf{0.324} \\
Electricity & 0.253 & \textbf{0.249} & 0.260 & \textbf{0.255} & 0.264 & \textbf{0.259} \\
Traffic & 0.264 & \textbf{0.257} & 0.276 & \textbf{0.269} & 0.295 & \textbf{0.289} \\
ILI & 0.846 & \textbf{0.822} & 0.833 & \textbf{0.815} & 1.041 & \textbf{0.997} \\
Weather & 0.264 & \textbf{0.262} & 0.263 & \textbf{0.262} & 0.300 & \textbf{0.296} \\
\bottomrule
\end{tabular}
\end{minipage}
\end{table}

Across PatchTST, Mamba, and DLinear, CARE consistently lowers errors on every dataset.
Averaged over eight benchmarks, relative reductions are 2.65\%, 3.30\%, and 3.31\% in MSE
and 1.74\%, 1.85\%, and 1.94\% in MAE for PatchTST, Mamba, and DLinear, respectively
(3.09\% and 1.84\% macro-averaged over the three backbones), indicating that the branch is
portable rather than tied to one architecture family.

\subsection{Input-Mode Comparison}
\label{subsec:input}
Section~\ref{secmethod} motivates feeding CARE only the aligned history $U=\mathcal{A}(x)$.
We compare four input maps $\Phi(\cdot)$ on PatchTST base forecaster on ETT and Weather:
(1)~$\Phi(U){=}U$;
(2)~$\Phi(U){=}Y^{\mathrm{base}}$;
(3)~$\Phi(U){=}[Y^{\mathrm{base}},U]$;
(4)~$\Phi(U){=}Y^{\mathrm{base}}-U$.
Table~\ref{tab:input} summarizes the results.

\begin{table}[t]
\caption{Input-mode comparison, averaged over
$T_p\in\{96,192,336,720\}$ per dataset.}
\label{tab:input}
\centering
\small
\setlength{\tabcolsep}{3pt}
\resizebox{\textwidth}{!}{%
\begin{tabular}{lcccccccccc}
\toprule
Input $\Phi(\cdot)$ &
\multicolumn{2}{c}{ETTh1} & \multicolumn{2}{c}{ETTh2} &
\multicolumn{2}{c}{ETTm1} & \multicolumn{2}{c}{ETTm2} &
\multicolumn{2}{c}{Weather} \\
\cmidrule(lr){2-3}\cmidrule(lr){4-5}\cmidrule(lr){6-7}\cmidrule(lr){8-9}\cmidrule(lr){10-11}
& MSE & MAE & MSE & MAE & MSE & MAE & MSE & MAE & MSE & MAE \\
\midrule
$U$ & \textbf{0.404} & \textbf{0.427} & \textbf{0.322} & \textbf{0.376} & \textbf{0.347} & \textbf{0.377} & \textbf{0.249} & \textbf{0.311} & \textbf{0.223} & \textbf{0.264} \\
$Y^{\mathrm{base}}$ & 0.412 & 0.434 & 0.328 & 0.381 & 0.350 & 0.379 & 0.255 & 0.316 & 0.231 & 0.266 \\
$[Y^{\mathrm{base}}, U]$ & 0.410 & 0.432 & 0.325 & 0.379 & 0.350 & 0.378 & 0.253 & \textbf{0.311} & 0.228 & 0.265 \\
$Y^{\mathrm{base}} - U$ & 0.409 & 0.431 & 0.326 & 0.380 & 0.350 & 0.381 & 0.252 & 0.314 & 0.229 & 0.267 \\
\bottomrule
\end{tabular}}%
\end{table}

Using only length-matched history achieves the lowest error on every dataset, suggesting
that the first convolutional stage should read calibrated past dynamics.
$\Phi(U){=}Y^{\mathrm{base}}$ is worst or tied for worst on most dataset--metric pairs.
Feeding only $Y^{\mathrm{base}}$ may lead the branch to re-fit or amplify the preliminary
trajectory instead of mining residual structure from the aligned past dynamics $U$.
The concatenation $\Phi(U){=}[Y^{\mathrm{base}},U]$ may partially restore $U$, yet may
still dilute useful information with the same preliminary signal.
The difference map $\Phi(U){=}Y^{\mathrm{base}}-U$ may further mix forecast-level
semantics with resampled historical dynamics, and shallow convolutions may then struggle
to extract stable correction cues from that input.

\subsection{Risk-Gating Effectiveness}
\label{subsec:riskgate}
On ETT and Weather with the PatchTST base forecaster, we test whether the gate $G$
tracks where the base model errs on the test set.
At each coordinate $(t,c)$ we compare $G$ to the absolute base-model error and report
three point-level diagnostics averaged over horizons in Table~\ref{tab:risk}.
AUROC assesses how well $G$ ranks high-error coordinates before low-error ones.
It targets the accuracy of this relative difficulty ordering.
To test whether $G$ captures structure in the data beyond its marginal distribution, we
define $\Delta\mathrm{AUROC}=\mathrm{AUROC}(G)-\mathrm{AUROC}(G_{\mathrm{perm}})$,
where $G_{\mathrm{perm}}$ randomly permutes gate values across points while preserving
the same marginal histogram.
A positive $\Delta$AUROC indicates that the ranking ability of $G$ exceeds random
reassignment and that $G$ encodes information beyond the raw value distribution alone.
Finally, $r_G$ is the ratio of mean test MAE in the highest versus lowest tertile of
$G$, and values above one mean larger gates co-occur with larger realized errors.

\begin{table}[t]
\caption{Risk-gating diagnostics on the test set (PatchTST + CARE), averaged over
$T_p\in\{96,192,336,720\}$ per dataset.}
\label{tab:risk}
\centering
\small
\setlength{\tabcolsep}{5pt}
\begin{tabular}{lccc}
\toprule
Dataset & AUROC & $\Delta$AUROC & $r_G$ \\
\midrule
ETTh1 & 0.7319 & 0.2319 & 2.2482 \\
ETTh2 & 0.7280 & 0.2281 & 2.1629 \\
ETTm1 & 0.7651 & 0.2652 & 2.7810 \\
ETTm2 & 0.7482 & 0.2481 & 2.1442 \\
Weather & 0.7926 & 0.2926 & 3.7762 \\
\bottomrule
\end{tabular}
\end{table}

Table~\ref{tab:risk} shows a consistent pattern across the five benchmarks.
AUROC lies between $0.73$ and $0.77$ on the four ETT splits and reaches $0.79$ on
Weather, indicating clearly above-chance ranking of high-error
coordinates.
$\Delta$AUROC is strictly positive on every dataset, from $0.23$ on ETTh2 to $0.29$
on Weather, so the learned gate is more informative than a permuted control with the
same marginal distribution.
$r_G$ exceeds $2$ throughout, from $2.14$ on ETTm2 to $3.78$ on Weather, so larger
gates co-occur with higher realized MAE on average.
Together, the three diagnostics support the intended role of $G$ as a soft indicator
of relative forecast difficulty, while we do not interpret them as proof of
calibrated probabilistic uncertainty.

\subsection{Ablation Study}
\label{subsec:ablation}
We ablate components of CARE and the training objective on ETT and
Weather with PatchTST base forecaster, averaging over $T_p\in\{96,192,336,720\}$.
Rows remove, in turn: scale-aware clipping ($\kappa{\to}\infty$), the gate loss
$\mathcal{L}_{\mathrm{risk}}$, the residual loss $\mathcal{L}_{\delta}$, the anchor
$\mathcal{L}_{\mathrm{anchor}}$, and the entire CARE branch (base forecaster only).
Table~\ref{tab:ablation} summarizes the ablations.

On average, across all datasets in Table~\ref{tab:ablation}, the full model achieves
the lowest test errors. Removing scale-aware clipping substantially
increases error, and dropping $\mathcal{L}_{\mathrm{risk}}$ also degrades metrics, indicating that
correction stability is disrupted.
Without $\mathcal{L}_{\delta}$ or $\mathcal{L}_{\mathrm{anchor}}$, accuracy declines
as well, which may reflect that under these training objectives the model ignores or
bypasses the base forecaster in order to reduce the loss.
We also sweep the clipping coefficient $\kappa\in\{0.1,0.2,0.5,1.0\}$
(Appendix Table 2).
At $\kappa{=}1.0$, error is worse than for smaller values, yet remains far below
\emph{w/o clip.}, which suggests that retaining some clipping remains helpful even when
the cap is loose.
For $\kappa\in\{0.1,0.2,0.5\}$, performance is highly similar. The mean coefficient of
variation of test MSE across prediction horizons averages about $2.1\%$, suggesting
limited sensitivity to $\kappa$ within this range.
The best average validation MSE occurs at $\kappa{=}0.1$, which we therefore adopt as
the default in all main experiments.

\begin{table}[t]
\caption{Ablation study (PatchTST; test MSE/MAE averaged over
$T_p\in\{96,192,336,720\}$ per dataset).}
\label{tab:ablation}
\centering
\small
\setlength{\tabcolsep}{3pt}
\resizebox{\textwidth}{!}{%
\begin{tabular}{lcccccccccc}
\toprule
Configuration &
\multicolumn{2}{c}{ETTh1} & \multicolumn{2}{c}{ETTh2} &
\multicolumn{2}{c}{ETTm1} & \multicolumn{2}{c}{ETTm2} &
\multicolumn{2}{c}{Weather} \\
\cmidrule(lr){2-3}\cmidrule(lr){4-5}\cmidrule(lr){6-7}\cmidrule(lr){8-9}\cmidrule(lr){10-11}
& MSE & MAE & MSE & MAE & MSE & MAE & MSE & MAE & MSE & MAE \\
\midrule
Full & \textbf{0.404} & \textbf{0.427} & \textbf{0.322} & \textbf{0.376} & \textbf{0.347} & \textbf{0.377} & \textbf{0.249} & \textbf{0.311} & \textbf{0.223} & \textbf{0.264} \\
w/o clip. & 0.417 & 0.440 & 0.529 & 0.541 & 0.364 & 0.393 & 0.727 & 0.541 & 0.228 & 0.271 \\
w/o $\mathcal{L}_{\mathrm{risk}}$ & 0.405 & \textbf{0.427} & 0.327 & 0.380 & 0.354 & 0.382 & 0.255 & 0.315 & 0.226 & 0.266 \\
w/o $\mathcal{L}_{\delta}$ & 0.406 & 0.429 & 0.327 & 0.381 & 0.353 & 0.382 & 0.253 & 0.314 & 0.226 & 0.266 \\
w/o $\mathcal{L}_{\mathrm{anchor}}$ & \textbf{0.404} & \textbf{0.427} & 0.327 & 0.381 & 0.352 & 0.381 & 0.256 & 0.315 & 0.227 & 0.267 \\
w/o CARE & 0.413 & 0.434 & 0.331 & 0.381 & 0.353 & 0.382 & 0.256 & 0.317 & 0.226 & \textbf{0.264} \\
\bottomrule
\end{tabular}}%
\end{table}

\subsection{Efficiency Analysis}
\label{subsec:efficiency}
\begin{figure}[H]
    \centering
    \includegraphics[width=\linewidth]{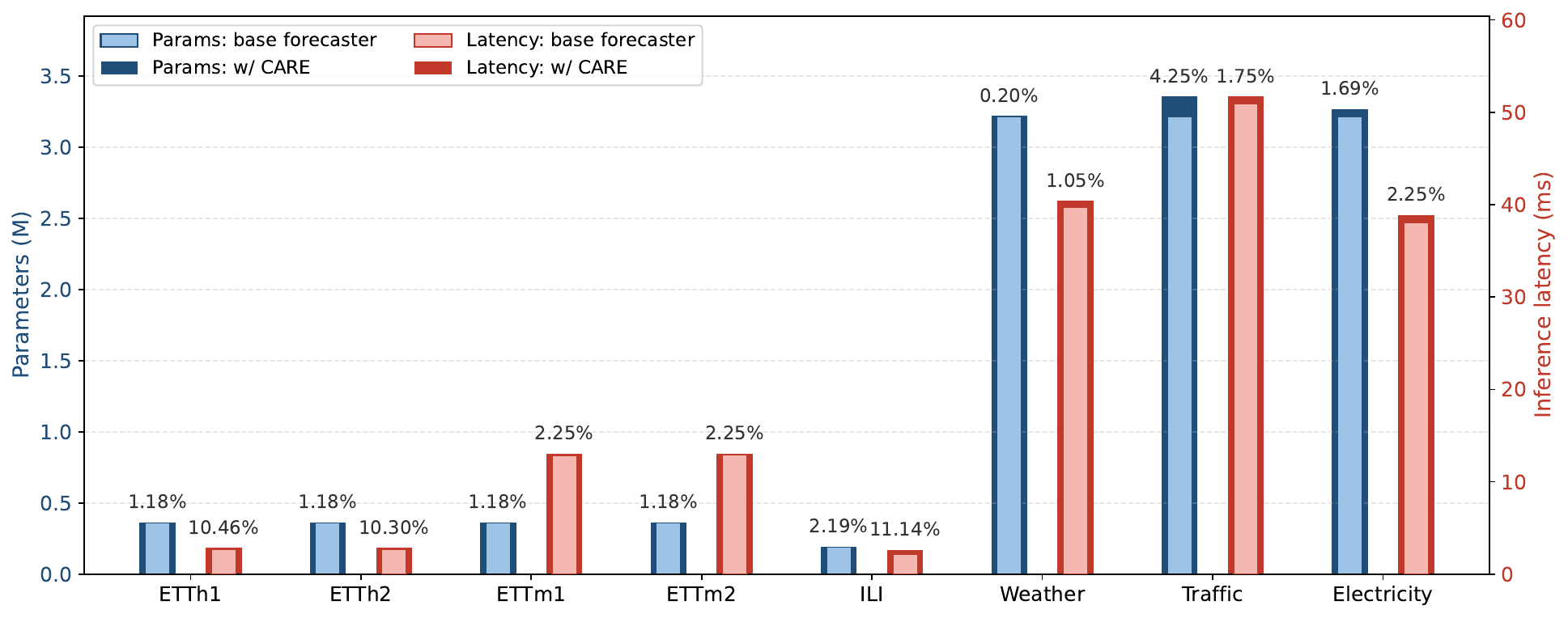}
    \caption{Parameter and inference-time overhead of CARE with PatchTST base forecaster.
    Blue bars show trainable parameters in millions.
    Red bars show mean forward latency in milliseconds.
    Light bars are the base forecaster only and dark bars are the full model with CARE.
    Percentages give CARE's share of the full model.}
    \label{fig:efficiency}
\end{figure}

We profile trainable parameters and inference latency for the PatchTST
base forecaster with and without CARE on ETT, ILI, Weather, Traffic, and Electricity
under the same horizon settings as in Section~\ref{subsec:impl}.
For latency, we warm up the GPU with 30 forward passes, then run 100 additional
forward passes and report the mean wall-clock time per pass in milliseconds.
Parameter totals count all trainable weights in $f_{\theta}$, including the full
base forecaster pipeline, and in CARE.
Figure~\ref{fig:efficiency} summarizes both metrics.
Blue bars show parameters on the left axis and red bars show inference time on the
right axis.
Light bars correspond to base forecaster only and dark bars to the full model with CARE.
The percentage above each bar is CARE's share of the full value.

The parameter overhead of CARE stays small on every benchmark.
On Weather it is $0.20\%$, on the ETT splits $1.18\%$ according to Appendix
Table 2, on Electricity $1.69\%$, on ILI $2.19\%$, and on Traffic
at most $4.25\%$, where the correction branch is widest.
Inference follows a similar pattern in absolute terms on large multivariate sets.
Weather rises from $39.7$\,ms to $40.4$\,ms, a relative increase of $1.05\%$.
Traffic rises from $50.9$\,ms to $51.7$\,ms with a relative increase of $1.75\%$, and
Electricity from $38.1$\,ms to $38.9$\,ms with a relative increase of $2.25\%$.
On ETTm1 and ETTm2 the base forecaster already costs about $12.9$\,ms per pass, and
the full model adds only about $0.2$\,ms, or about $2.25\%$ relative to that baseline.
On ETTh1, ETTh2, and ILI the relative latency share is about $10\%$ to $11\%$ because
the $f_{\theta}$ forward pass is very short, about $2.2$ to $2.7$\,ms.
The extra about $0.3$\,ms from CARE is small in absolute terms but shows up as a
larger percentage.
In summary, the extra parameters and latency added by CARE are limited relative to
the base forecaster.

\section{Conclusion and Future Work}
\label{sec:conclusion}
In this paper, we propose CARE, a lightweight plug-in for long-horizon forecasting
that targets a key limitation of mainstream deterministic methods, namely that they
typically output only a single trajectory and offer little explicit control over
correction.
CARE reads length-matched historical input through a shallow convolutional branch and,
with per-coordinate risk-score gating, supplies bounded residual corrections to the
base forecaster while preserving most of the base model's capacity.
CARE therefore focuses on residual structure that the base predictor tends to overlook.
Across benchmarks, attaching CARE to PatchTST, Mamba, and DLinear
consistently improves MSE and MAE yet adds only a small number of trainable parameters
and modest forward latency.
Future work includes extending CARE to probabilistic forecasting frameworks,
exploring mechanisms to calibrate the gate weights, and validating generalization
under richer covariate information and large-scale practical settings.

\begin{credits}
    \subsubsection{\ackname}
    This research was supported by the Science and Technology Project of the Ministry of Agriculture and Rural Affairs, Beijing Municipal Science and Technology Project (No. Z251100007525010), and the Fundamental Research Funds for the Central Universities (No. FRF-BD-26-033).
    
    \subsubsection{\discintname}
    The authors have no competing interests to declare that are relevant to the content of this article.
\end{credits}
%
% Bibliography
%
% BibTeX users should specify bibliography style splncs04.
% References will then be sorted and formatted in the correct style.
%
% \bibliographystyle{splncs04}
% \bibliography{mybibliography}
%

\end{document}